# ORDINAL SCALE TRAFFIC CONGESTION CLASSIFICATION WITH MULTI-MODAL VISION-LANGUAGE AND MOTION ANALYSIS

*Yu-Hsuan Lin*[1]

Dept. of Artificial Intelligence Technology and Application,
Feng Chia University, Taichung, Taiwan
E-mail: D1278799@o365.fcu.edu.tw

**ABSTRACT**

Accurate traffic congestion classification is essential for intelligent transportation systems and real-time urban traffic management. This paper presents a multimodal framework combining open-vocabulary visual-language reasoning (CLIP), object detection (YOLO-World), and motion analysis via MOG2-based background subtraction. The system predicts congestion levels on an ordinal scale from 1 (free flow) to 5 (severe congestion), enabling semantically aligned and temporally consistent classification. To enhance interpretability, we incorporate motion-based confidence weighting and generate annotated visual outputs. Experimental results show the model achieves 76.7% accuracy, an F1-score of 0.752, and a Quadratic Weighted Kappa (QWK) of 0.684, significantly outperforming unimodal baselines. These results demonstrate the framework's effectiveness in preserving ordinal structure and leveraging visual-language and motion modalities. Future enhancements include incorporating vehicle sizing and refined density metrics.

## 1. INTRODUCTION

In the context of urbanization and the continuous increase in the number of vehicles, traffic congestion has grown into a major challenge for urban management. It not only poses a threat to the overall traffic environment but also adversely affects the quality of daily life for residents. There is a need to develop effective traffic management systems to alleviate congestion. A key feature of these systems is their capacity to classify congestion levels in real-time and with high accuracy, providing practical information for urban planning, adaptive signal control, and intelligent transportation systems (ITS) [1]. Despite their significance, conventional techniques for classifying congestion often struggle to address the complexity of real-world scenarios. These methods usually rely on a single data source, such as visual features captured by CCTV cameras, or depend heavily on manual labeling. However, factors such as changing lighting conditions, vehicle occlusion, and background motion dynamically affect traffic scenes, which are inherently complex. Such variability and noise can degrade system performance [2].

To address these drawbacks, this paper proposes an innovative ordinal scale traffic congestion classification framework that integrates multimodal visual-language information with dynamic analysis. The proposed method utilizes three complementary data modalities: the first is visual features extracted from video frames; and the second is semantic understanding for zero-shot classification through language prompts; the third is temporal dynamics derived from vehicle motion analysis. Particularly, we use the YOLO (You Only Look Once) object detection model to identify and count vehicles in each frame, providing an accurate metric for traffic density [3,4,5]. At the same time, the CLIP (Contrastive Language-Image Pretraining) model is used to perform zero-shot classification, aligning visual scenes with predefined text prompts (for example, from "free-flowing traffic" to "severe congestion"). Background subtraction techniques are used to analyze motion-related metrics, such as motion percentage and stability, which represent the speed and behavior of vehicles over time to capture the dynamic characteristics of traffic flow [6,7]. Our framework overcomes the limitations of conventional unimodal approaches by using multimodal data to classify traffic congestion levels more precisely and comprehensively [8].

The key innovation of the research presented consists of ordinal scale classification, dividing traffic congestion into five levels (Level 1 to Level 5). Compared to traditional binary or nominal classification methods, this approach can more accurately reflect the severity of congestion, providing a more detailed representation of traffic dynamics, and is particularly suitable for precise applications such as congestion prediction and signal optimization. Our experimental framework employs a time segmentation method to process image data and uses multimodal models for inference, aggregating frame-level predictions into period-level congestion ratings.

## 2. RELATED WORK

Recent studies in traffic congestion analysis have increasingly focused on leveraging multimodal approaches to enhance classification robustness and adaptability. With the growing deployment of urban surveillance and AI-driven infrastructure, researchers have explored combining semantic understanding, visual detection, and motion modeling to better capture the complexity of real-world traffic scenes. This section highlights two core areas underpinning our framework: open-vocabulary object recognition, which enables semantic-level inference through image-text alignment, and background subtraction, which extracts temporal motion cues for dynamic scene analysis.

### 2.1. Open Vocabulary Object Recognition (OVD)

In recent years, open-vocabulary object detection has achieved significant advancements in the field of object recognition. Its core capability lies in effectively identifying previously unseen object categories under zero-shot learning scenarios. Unlike conventional object detection models such as YOLO, SSD, or Faster R-CNN, which rely heavily on large volumes of annotated training data [9,10,11], open-vocabulary approaches leverage semantic embedding models such as CLIP to associate textual descriptions with visual features, enabling cross-modal matching [12,13,14,15]. This makes them highly adaptable to diverse application scenarios, particularly in aerial imagery, where object variability and viewpoint changes are prominent challenges.

By learning a shared embedding space through semantic models, visual feature representations $f_{img} \in \mathbb{R}^d$ and textual semantic features $f_{text} \in \mathbb{R}^d$ are projected into the same latent space. The similarity between an image and a textual prompt is then computed through an inner product, as expressed as Eq. (1).

$$S = f_{img} \cdot f_{text} \quad (1)$$

As illustrated in Fig. 1. The system framework projects both image and textual features into a common embedding space, thereby enabling matching through the computation of feature similarity. This integrated representation supports efficient and precise cross-modal recognition.

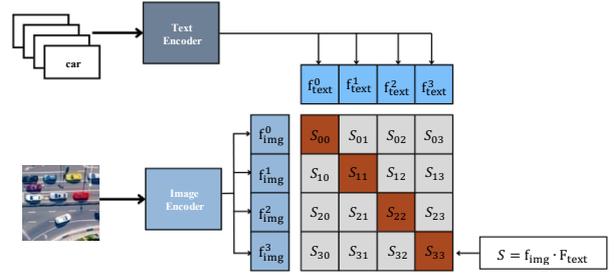

Fig. 1. Framework for Feature Matching between Image and Text Representations

The image is processed through an image encoder to generate features $f_{img}$, while multiple label texts are encoded through a text encoder to produce text features $f_{text}^i$. These features are subjected to an inner product operation to generate a similarity matrix $S$, wherein the text corresponding to the highest similarity score is assigned as the target category for the image. In this context, $S$ represents the matching scores between the image and text. For multi-object detection scenarios, the similarity matrix for the features of all candidate boxes $\{f_{img}^i\}_{i=1}^N$ and target descriptions $\{f_{text}^i\}_{j=1}^M$ is formulated as shown as Eq. (2).

$$S_{ij} = f_{img}^i \cdot f_{text}^j \quad (2)$$

Here, $N$ denotes the number of candidate boxes, while $M$ represents the number of target descriptions. The open-vocabulary model plays a pivotal role in enabling accurate multi-object localization and classification within an image by optimizing the matching scores between the features [9]. Through the application of these techniques, this study introduces a real-time open-vocabulary detection approach for the initial localization and classification of objects in dynamic scenes. This approach significantly enhances recognition accuracy and improves the model's generalization ability across a wide range of scenarios.

### 2.2. Background Subtraction

Background subtraction is widely adopted in video analysis for isolating moving objects from static backgrounds, making it essential for traffic monitoring applications where vehicle motion indicates congestion [16]. The Mixture of Gaussians (MOG2) algorithm offers robustness to gradual illumination changes and is commonly used in traffic video analysis [17]. MOG2 models each pixel's intensity as a mixture of Gaussian distributions, updated over time to adapt to environmental changes, and includes shadow detection to reduce false positives in foreground segmentation.

In the proposed system, foreground masks are extracted from video frames using MOG2. These masks are thresholded and processed through morphological operations (e.g., opening and closing with an elliptical kernel) to reduce noise and enhance object contours. The processed masks enable computation of motion-based metrics, including motion coverage (percentage of active pixels), motion intensity, and contour density [18].

These metrics, derived through post-processing steps, support analysis of motion patterns across frames. For example, motion stability and trend (e.g., increasing, decreasing, or stable motion) are computed from the history of foreground masks, offering insights into traffic flow dynamics that improve congestion classification [19]. In this context, background subtraction complements vehicle detection by providing temporal context for a more comprehensive interpretation of traffic behavior.

While recent advances in background subtraction have focused on deep learning to improve robustness in challenging scenarios such as occlusions or sudden lighting changes [20], traditional methods like MOG2 remain a computationally efficient and effective choice for real-time applications, especially when integrated with complementary vision techniques. Therefore, our framework opts for a lightweight MOG2-based approach, leveraging its speed and reliability. Moreover, motion-based metrics from MOG2 provide valuable temporal context for congestion estimation, complementing vehicle count-based estimates from object detection.

## 3. METHODOLOGY

The proposed traffic congestion classification framework integrates the CLIP vision-language model with YOLO object detection and motion analysis into a modular system comprising five components: (1) Image Input & Preprocessing, (2) CLIP-Based Visual Feature Extraction, (3) YOLO Vehicle Detection, (4) Background Subtraction and Motion Analysis, and (5) Ordinal Congestion Reasoning and Classification. By leveraging CLIP's zero-shot classification capability with predefined textual prompts representing congestion states, the system generalizes across scenes without retraining. YOLO provides quantitative measures of vehicle count and density, while MOG2-based background subtraction extracts motion features to refine predictions. The ordinal reasoning module fuses these inputs for robust classification. An overview of the system architecture is shown in Fig. 2.

### 3.1. Image Input & Preprocessing
Each input frame is resized (e.g., to 224×224 pixels) and converted from BGR to RGB format to conform with CLIP's image encoder requirements. The processed frames are also used by downstream modules for vehicle detection and motion segmentation.

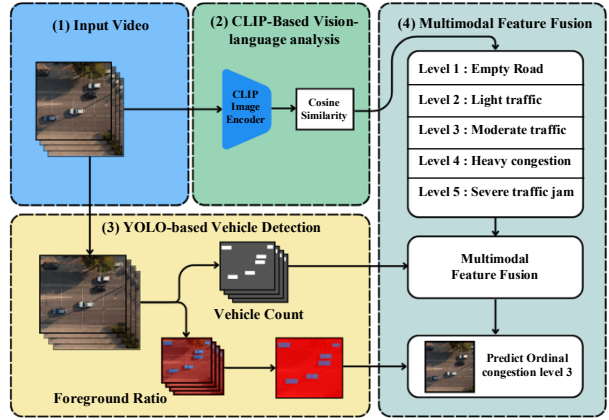

Fig. 2. System architecture for ordinal-scale traffic congestion classification.

### 3.2. CLIP Visual Feature Extraction
To classify traffic conditions from images without needing labeled training data, we leverage CLIP's zero-shot learning capabilities. This approach allows us to compare visual frames directly to predefined text descriptions (e.g., "light traffic", "heavy congestion") using semantic similarity.

Each video frame is first preprocessed and passed through CLIP's image encoder to produce a visual embedding vector $f_v \in \mathbb{R}^d$, which captures the semantic content of the frame. In parallel, a list of text prompts describing each congestion level is passed through CLIP's text encoder to produce a set of textual embedding vectors $\{f_t^1, f_t^2, \cdots f_t^5\}$, as corresponding to one of five congestion levels. To determine how well the image matches each textual description, we compute the cosine similarity between the image embedding and each text embedding as Eq. (3).

$$\text{Similarity}(f_v, f_t) = \cos(f_v, f_t) = \frac{f_v \cdot f_t}{\|f_v\| \|f_t\|} \quad (3)$$

This yields a similarity vector $s = [s1, s2, \cdots, s5]$, where each value $s_i$ indicates how strongly the frame matches the congestion level. The highest score in this vector determines the predicted congestion category.

### 3.3. YOLO Vehicle Detection
We utilize a YOLO-based model ("YOLO World") to detect vehicles at each frame $t$, producing a set of bounding boxes $B_t = \{b_1, \cdots, b_{N_t}\}$. The number of detected vehicles $N_t = |B_t|$ serves as a proxy for traffic density. Each bounding box $b_i$ contains position and confidence data. While the number of detected vehicles is used to estimate traffic density, modeling of vehicle size or bounding box area has not yet been implemented.

### 3.4. Background Subtraction & Motion Analysis
To extract motion-related features, we apply the MOG2 background subtraction algorithm to each frame, generating a binary foreground mask. Morphological operations (e.g., opening and closing with an elliptical kernel) are used to reduce noise and refine object contours. Motion stability is estimated over a temporal window by computing heuristic variance metrics based on recent frame activity, as shown in Eq. (4).

$$S_t = \frac{1}{1 + \text{Var}(\mu_{t-w}, \cdots \mu_t)} \quad (4)$$

Here, $\mu_t$ denotes the ratio of foreground (motion) pixels to total pixels in frame $t$. The window size $w$ is empirically set to 15 frames, balancing responsiveness, and stability. Although no explicit tracking or velocity estimation is performed, motion features serve as an auxiliary signal to adjust prediction confidence. The current MOG2-based motion module lacks object-level tracking or velocity estimation, which will be explored in future work.

### 3.5. Ordinal Congestion Reasoning and Classification Strategy

For each frame, the system fuses multiple features by CLIP similarity scores and contextual motion features to estimate a congestion level $L_t \in \{1, \ldots, 5\}$. Segment-level predictions are obtained by aggregating frame-level estimates using a weighted average or median, followed by temporal smoothing as Eq. (5).

$$L_{smoothed} = \text{round}\left(\frac{1}{w} \sum_{i=t-w/2}^{t+w/2} L_i\right) \quad (5)$$

The final output includes a predicted congestion level and a corresponding confidence score derived from CLIP similarity. The output follows an ordinal structure, where $L \in \{1, \ldots, 5\}$, reflects increasing levels of congestion.

By preserving the inherent order of traffic states, the ordinal structure enhances interpretability and enables finer granularity in monitoring traffic trends over time. Unlike traditional classifiers that treat congestion states as independent categories, our approach preserves the inherent ordering of traffic conditions, improving temporal stability and reducing abrupt class transitions. Each frame is passed through the CLIP encoder to generate a visual feature vector $f_v$, which is compared against five textual embeddings representing congestion levels. Cosine similarity yields a score vector $s = [s_1, \cdots, s_5]$, and the label with the highest score is selected as the prediction as Eq. (6).

$$\hat{L}_t^{adj} = \underset{i}{\text{argmax}}\ s_i \quad (6)$$

To incorporate temporal and contextual nuances, these scores are refined using a scalar adjustment factor based on motion patterns. We define an adjusted score vector $\hat{s}_i = s_i \cdot \alpha_i$, where $\alpha_i$ encodes heuristic adjustments based on foreground activity and contour presence. For example, frames with minimal motion and consistent object contours favor higher congestion levels. In our current implementation, $\alpha_i$ is a rule-based scalar. For instance, if motion coverage is below 5% over a span of 15 frames, $\alpha_i$ is increased to 1.2 to amplify congestion likelihood. Conversely, stable high motion results in $\alpha_i = 0.8$, indicating free-flow conditions. Final predictions are computed at the segment level by aggregating frame-level estimates using confidence-weighted averaging as described in Eq. (7).

$$L_{seg} = \text{round}\left(\frac{\sum_{t \in \mathcal{T}} c_t \cdot \hat{L}_t^{adj}}{\sum_{t \in \mathcal{T}} c_t}\right) \quad (7)$$

Where $c_t = \max \hat{s}_i$ and $\mathcal{T}$ denotes the set of frames in the segment. A smoothing filter is then applied across adjacent segments to suppress outlier effects. This classification strategy, grounded in ordinal-aware ranking, multimodal fusion, and temporal consistency, enables the system to deliver precise, interpretable, and real-time assessments of traffic congestion across diverse urban scenarios.

## 4. EXPERIMENTAL SETUP

### 4.1. Data Description

We conducted experiments on urban traffic video sequences recorded under different congestion conditions: free flow, moderate, and severe. Videos were manually segmented into 100-frame clips, with each segment labeled based on congestion level from 1 (free flow) to 5 (severe congestion). Although the dataset is limited in scale, it was selected to represent clear distinctions between congestion levels to evaluate our ordinal classification framework.

To ensure temporal consistency and reduce label noise, predictions were evaluated at the segment level rather than per frame. Ground-truth labels were established based on visual inspection of traffic density and flow within each segment.

### 4.2. Implementation Details

The proposed framework was developed in Python and implemented using PyTorch to facilitate modular integration of the vision-language and detection models. For semantic feature extraction, we employed the CLIP model by the API from OpenAI, allowing for zero-shot inference with congestion-level text prompts. Vehicle detection was carried out using a pretrained YOLO-World model, optimized for object recognition in traffic scenarios.

Motion analysis was conducted using a Gaussian Mixture Model (GMM)-based background subtraction technique, specifically an implementation of the MOG2 algorithm. This approach was selected for its efficiency, adaptability to gradual illumination changes, and suitability for real-time foreground segmentation without requiring extensive computational resources. The extracted binary foreground masks were further refined using standard morphological operations to enhance contour clarity and suppress noise.

All experiments were conducted on a high-performance workstation configured with an Intel Core i7-12700 CPU, an NVIDIA RTX 3080 Ti GPU (12GB VRAM), 64 GB of RAM, and 2.5 TB of SSD storage, running Windows 11 (version 24H2).

## 5. RESULT

### 5.1. Quantitative and Qualitative Results

This section evaluates our multimodal traffic congestion classification system against three unimodal baselines: CLIP only, YOLO only, and MOG2 only. Performance is assessed using classification accuracy, mean absolute error (MAE), F1 score, Quadratic Weighted Kappa (QWK), and average inference time per segment, as presented in Table 1.

Table 1. Performance comparison of the proposed model and baselines.

| Model | Accuracy | MAE | F1 | QWK | Time (s) |
|---|---|---|---|---|---|
| CLIP Only | 0.422 | 0.656 | 0.282 | 0.212 | 1.90 |
| YOLO Only | 0.456 | 0.772 | 0.410 | 0.251 | 8.42 |
| MOG2 Only | 0.022 | 1.817 | 0.003 | 0.000 | 2.15 |
| **Proposed Method** | **0.767** | **0.233** | **0.752** | **0.684** | **14.03** |

The proposed method outperforms all baselines, achieving the highest accuracy (76.7%), lowest MAE (0.233), and best F1 score (0.752), indicating robust agreement with ground-truth labels and balanced performance across congestion levels. Its QWK of 0.684 highlights superior ordinal consistency, minimizing semantically disruptive misclassifications. While the method incurs a higher computational cost (14.03 seconds per segment), this trade-off is justified by significant gains in interpretability and prediction stability.

In contrast, the MOG2 only baseline performs poorly, with near-zero metrics, confirming that motion features alone are insufficient. The CLIP only and YOLO only models yield moderate results, limited by the absence of temporal dynamics and semantic context, respectively. The integration of semantic, spatial, and motion cues in our approach proves essential for accurate congestion classification.

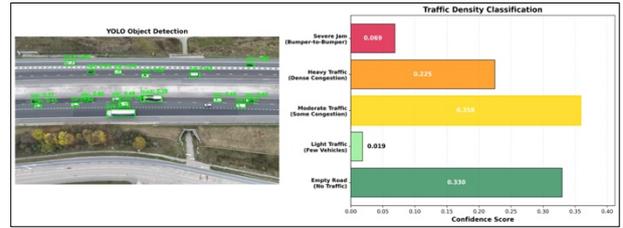

Fig. 3. YOLO Object Detection and Traffic Density Classification.

Qualitative analysis further underscores the system's practical utility. The system's output is visualized in Fig. 3. Which shows an aerial highway image with vehicles detected by YOLO, marked with green bounding boxes, alongside a bar chart displaying confidence scores for traffic density: Moderate Traffic (0.358), Empty Road (0.330), Heavy Traffic (0.225), Severe Jam (0.069), Light Traffic (0.019). This dual representation reflects the algorithm's assessment of mixed traffic conditions, where the high confidence in both severe and light congestion aligns with the observed vehicle spacing, demonstrating effective real-world application.

### 5.2. Theoretical Justification for Ordinal Classification

The ordinal-scale classification framework is adopted due to the inherently ordered nature of traffic congestion states, ranging from "free-flow" to "severe congestion." Unlike multi-class classification, which treats classes independently, ordinal classification preserves semantic proximity between adjacent levels (e.g., Level 2 is closer to Level 3 than to Level 5). This aligns with human perception and supports real-world applications, such as adaptive signal control, where distinguishing subtle congestion differences is critical.

A five-level scale, informed by empirical observations and conventional traffic grading systems (e.g., LOS A–F standards, mapped to levels 1–5), balances interpretability and resolution. It captures meaningful variations while remaining robust against labeling noise and classifier ambiguity. Regression approaches, though suitable for continuity, lack the discrete outputs required by intelligent transportation system (ITS) interfaces. Our ordinal structure, by contrast, leverages inherent order to reduce misclassification impact, as evidenced by the high QWK score (0.684), ensuring consistent ordinal relationships.

### 5.3. Discussion

The results demonstrate that our multimodal approach effectively leverages complementary features semantic context from CLIP, spatial detection from YOLO, and temporal motion from MOG2 to achieve superior congestion classification. The high accuracy (76.7%) and QWK (0.684) indicate not only predictive power but also the model's ability to respect the ordinal nature of traffic states, a critical factor for ITS applications. The qualitative evidence in Fig 3 further validates the system's interpretability, showing how it handles real-world ambiguity in traffic scenes. However, the reliance on frame-wise motion analysis suggests potential limitations in dynamic or low-visibility conditions, as noted in the limitations. These findings lay a foundation for future enhancements, particularly in real-time deployment and robustness across diverse environments.

## 6. CONCLUSION

This study introduces a novel multimodal framework for ordinal-scale traffic congestion classification, integrating vision-language modeling (CLIP), object detection (YOLO), and motion analysis (MOG2). By fusing semantic, spatial, and temporal cues, the system achieves robust congestion estimation, outperforming unimodal baselines with an accuracy of 76.7%, an F1 score of 0.752, and a Quadratic Weighted Kappa (QWK) of 0.684, reflecting strong ordinal consistency. The adoption of a five-level ordinal scheme enhances interpretability and aligns with practical needs, such as adaptive signal control, by preserving semantic granularity.

Despite these advances, limitations include the absence of object-level tracking and untested performance under adverse conditions like nighttime or low visibility. Future work will focus on integrating trajectory-based motion analysis, developing adaptive ordinal scaling for varied traffic contexts (e.g., rush hour vs. off-peak), and evaluating scalability and latency for real-time edge deployment. These improvements aim to bolster precision, temporal insight, and applicability in intelligent transportation systems.